\pdfoutput=1

\documentclass[11pt]{article}

\usepackage[]{emnlp2021}

\usepackage{times}
\usepackage{latexsym}
\usepackage{comment}
\usepackage[T1]{fontenc}

\usepackage[utf8]{inputenc}

\usepackage{microtype}

\usepackage{lipsum}
\usepackage{graphicx} 

\usepackage{cleveref}
\crefname{section}{§}{§§}
\Crefname{section}{§}{§§}

\title{Exploiting Twitter as Source of Large Corpora of Weakly Similar Pairs for Semantic Sentence Embeddings}

\author{Marco Di Giovanni \\
  Politecnico di Milano \\
  Università di Bologna \\
  \texttt{marco.digiovanni@polimi.it} \\\And
  Marco Brambilla \\
  Politecnico di Milano \\
  \texttt{marco.brambilla@polimi.it} \\}

\begin{document}
\maketitle
\begin{abstract}
Semantic sentence embeddings are usually supervisedly built minimizing distances between pairs of embeddings of sentences labelled as semantically similar by annotators. Since big labelled datasets are rare, in particular for non-English languages, and expensive, recent studies focus on unsupervised approaches that require not-paired input sentences. 
We instead propose a language-independent approach to build large datasets of pairs of informal texts weakly similar, without manual human effort, exploiting Twitter's intrinsic powerful signals of relatedness: replies and quotes of tweets. 
We use the collected pairs to train a Transformer model with triplet-like structures, and we test the generated embeddings on Twitter NLP similarity tasks (PIT and TURL) and STSb. 
We also introduce four new sentence ranking evaluation benchmarks of informal texts, carefully extracted from the initial collections of tweets, proving not only that our best model learns classical Semantic Textual Similarity, but also excels on tasks where pairs of sentences are not exact paraphrases. 
Ablation studies reveal how increasing the corpus size influences positively the results, even at 2M samples, suggesting that bigger collections of Tweets still do not contain redundant information about semantic similarities.\footnote{Code available at \url{https://github.com/marco-digio/Twitter4SSE}}
\end{abstract}

\section{Introduction and Related Work}

Word-level embeddings techniques compute fixed-size vectors encoding semantics of words~\cite{NIPS2013_word2vec,pennington2014glove}, usually unsupervisedly trained from large textual corpora. It has always been more challenging to build high-quality sentences-level embeddings. 

Currently, best sentence-embeddings approaches are supervisedly trained using large labeled datasets~\cite{conneau-EtAl:2017:EMNLP2017,cer2018universal,reimers-2019-sentence-bert,chen-etal-2019-multi-task,du2020self,wieting-etal-2020-bilingual,huang2021disentangling}, such as NLI datasets~\cite{snli-emnlp2015,williams-etal-2018-broad} or paraphrase corpora~\cite{dolan-brockett-2005-automatically}. 
Round-trip translation has been also exploited, where semantically similar pairs of sentences are generated translating the non-English side of NMT pairs, as in ParaNMT~\cite{wieting-gimpel-2018-paranmt} and Opusparcus~\cite{creutz-2018-open}. 
However, large labeled datasets are rare and hard to collect, especially for non-English languages, due to the cost of manual labels, and there exist no convincing argument for why datasets from these tasks are preferred over other datasets~\cite{carlsson2021semantic}, even if their effectiveness on STS tasks is largely empirically tested. 

Therefore, recent works focus on unsupervised approaches~\cite{li-etal-2020-sentence,carlsson2021semantic,wang2021tsdae,giorgi2020declutr,logeswaran2018an}, where unlabeled datasets are exploited to increase the performance of models. 
These works use classical formal corpora such as OpenWebText~\cite{Gokaslan2019OpenWeb}, English Wikipedia, obtained through Wikiextractor~\cite{Wikiextractor2015}, or target datasets without labels, such as the previously mentioned NLI corpora. 

Instead, we propose a Twitter-based approach to collect large amounts of \textit{weak} parallel data: the obtained couples are not exact paraphrases like previously listed datasets, yet they encode an intrinsic powerful signal of relatedness. 
We test pairs of quote and quoted tweets, pairs of tweet and reply, pairs of co-quotes and pairs of co-replies. 
We hypothesize that quote and reply relationships are weak but useful links that can be exploited to supervisedly train a model generating high-quality sentence embeddings. 
This approach does not require manual annotation of texts and it can be expanded to other languages spoken on Twitter. 

We train models using triplet-like structures on the collected datasets and we evaluate the results on the standard STS benchmark~\cite{cer-etal-2017-semeval}, two Twitter NLP datasets~\cite{PIT,TURL} and four novel benchmarks. 

Our contributions are four-fold: we design an language-independent approach to collect big corpora of weak parallel data from Twitter; we fine-tune Transformer based models with triplet-like structures; we test the models on semantic similarity tasks, including four novel benchmarks; we perform ablation on training dataset, loss function, pre-trained initialization, corpus size and batch size. 

\section{Datasets}\label{sec:datasets}

We download the general Twitter Stream collected by the Archive Team
Twitter\footnote{\url{https://archive.org/details/twitterstream}}. We select English\footnote{\textit{English} tweets have been filtered accordingly to the ``lang" field provided by Twitter.} tweets posted in November and December 2020, the two most recent complete months up to now. They amount to about $27G$ of compressed data ($\sim 75M$ tweets).\footnote{We do not use the official Twitter API because it does not not guarantee a reproducible collections (Tweets and accounts are continuously removed or hidden due to Twitter policy or users' privacy settings). }
This temporal selection could introduce biases in the trained models since conversations on Twitter are highly related to daily events. 
We leave as future work the quantification and investigation of possible biases connected to the width of the temporal window, but we expect that a bigger window corresponds to a lower bias, thus a better overall performance. 

We collect four training datasets: the Quote Dataset, the Reply Dataset, the Co-quote Dataset and the Co-reply Dataset. 

The \textbf{Quote Dataset (Qt)} is the collection of all pairs of quotes and quoted tweets. 
A user can \textit{quote} a tweet by sharing it with a new comment (without the new comment, it is called \textit{retweet}). 
A user can also retweet a quote, but it cannot quote a retweet, thus a quote refers to an original tweet, a quote, or a reply. 
We generate \textit{positive} pairs of texts coupling the quoted texts with their quotes. 

The \textbf{Reply Dataset (Rp)} is the collection of all couples of replies and replied tweets. 
A user can reply to a tweet by posting a public comment under the tweet. 
A user can reply to tweets, quotes and other replies. 
It can retweet a reply, but it cannot reply to a retweet, as this will be automatically considered a reply to the original retweeted tweet. 
We generate \textit{positive} pairs of texts coupling tweets with their replies. 

The \textbf{Co-quote Dataset (CoQt)} and \textbf{Co-reply Dataset (CoRp)} are generated respectively from the Qt Dataset and the Rp Dataset, selecting as \textit{positive} pairs two quotes/replies of the same tweet. 

To avoid \textit{popularity-bias} we collect only one positive pair for each quoted/replied tweet in every dataset, otherwise viral tweets would have been over-represented in the corpora. 

We clean tweets by lowercasing the text, removing URLs and mentions, standardizing spaces and removing tweets shorter than $20$ characters to minimize generic texts (e.g., variations of "Congrats" are common replies, thus they can be usually associated to multiple original tweets). 
We randomly sample $250k$ positive pairs to train the models for each experiment, unless specified differently, to fairly compare the performances (in~\cref{sec:res} we investigate how the corpus size influences the results).  
We also train a model on the combination of all datasets (\textbf{all}), thus $1M$ text pairs. 

We show examples of pairs of texts from the four datasets in the Appendix.

\section{Approach}

We select triplet-like approaches to train a Tranformer model on our datasets. We extensively implement our models and experiments using sentence-transformers python library\footnote{\url{https://github.com/UKPLab/sentence-transformers}} and Huggingface~\cite{wolf-etal-2020-transformers}. 
Although the approach is model-independent, we select four Transfomer models~\cite{attention} as pre-trained initializations, currently being the most promising technique ($\sim 110M$ parameters):\\
\textbf{RoBERTa base}~\cite{roberta} is an improved pre-training of BERT-base architecture~\cite{devlin-etal-2019-bert}, to which we add a pooling operation: MEAN of tokens of last layer. Preliminary experiments of pooling operations, such as MAX and [CLS] token, obtained worse results;\\
\textbf{BERTweet base}~\cite{nguyen-etal-2020-bertweet} is a BERT-base model pre-trained using the same approach as RoBERTa on $850M$ English Tweets, outperforming previous SOTA on Tweet NLP tasks, to which we add a pooling operation: MEAN of tokens of last layer;\\
\textbf{Sentence BERT}~\cite{reimers-2019-sentence-bert} are BERT-base models trained with siamese or triplet approaches on NLI and STS data. We select two suggested base models from the full list of trained models: bert-base-nli-stsb-mean-tokens (S-BERT) and stsb-roberta-base (S-RoBERTa). \\

We test the two following loss functions:\\
\textbf{Triplet Loss (TLoss)}: given three texts (an anchor $a_i$, a positive text $p_i$ and a negative text $n_i$), we compute the text embeddings ($s_a$, $s_p$, $s_n$) with the same model and we minimize the following loss function:\\
$$max(||s_a-s_p||-||s_a-s_n||+\epsilon, 0)$$
For each pair of anchor and positive, we select a \textit{negative} text randomly picking a positive text of a different anchor (e.g., about the Quote dataset, anchors are quoted tweets, positive texts are quotes and the negative texts are quotes of different quoted tweets);\\
\textbf{Multiple Negative Loss (MNLoss)}~\cite{loss_multiple}: given a batch of positive pairs $(a_1, p_1), ..., (a_n, p_n)$, we assume that $(a_i, p_j)$ is a negative pair for $i\neq j$ (e.g., Quote Dataset: we assume that quotes cannot refer to any different quoted tweet). We minimize the negative log-likelihood for softmax normalized scores. We expect the performance to increase with increasing batch sizes, thus we set $n=50$, being the highest that fits in memory (see~\cref{sec:res} for more details). 

We train the models for 1 epoch\footnote{We briefly tested the training for two epochs in preliminary experiments, but we noticed no evident benefits. Moreover, increasing the number of epochs enhances the risk of overfitting the noise included in tweets since these texts are noisy and we do not perform validation.} with AdamW optimizer, learning rate $2 \times 10^{-5}$, linear scheduler with $10\%$ warm-up steps on a single NVIDIA Tesla P100. Training on $250k$ pairs of texts requires about $1$ hour, on $1M$ about $5$ hours. 

\section{Evaluation}

We evaluate the trained models on seven heterogeneous semantic textual similarity (STS) tasks: four novel benchmarks from Twitter, two well-known Twitter benchmarks and one classical STS task. 
We planned to test the models also on Twitter-based classification tasks, e.g., Tweeteval~\cite{barbieri-etal-2020-tweeteval}. However, the embeddings obtained from our approach are \textit{not} designed to transfer learning to other tasks, but they should mainly succeed on similarity tasks. 
A complete and detailed evaluation of our models on classification tasks is also not straightforward, since a classifier must be selected and trained on the top of our models, introducing further complexity to the study. We leave this analysis for future works. 

\subsection{Novel Twitter benchmarks}

We propose four novel benchmarks from the previously collected data. Tweets in these datasets are discarded from \textit{every} training set to avoid unfair comparisons. 
We frame these as ranking tasks and we pick normalized Discounted Cumulative Gain (nDCG) as metric~\cite{nDCG}\footnote{nDCG is a common ranking-quality metric obtained normalizing Discounted Cumulative Gain (DCG). The scores range from $0$ to $1$, the higher the better. Thus, $1$ represents a perfect ranking: the first ranked document is the most relevant one, the second ranked document is the second most relevant one, and so on. }. 
We propose these datasets to highlight that benchmark approaches are not able to detect similarities between related tweets, while they can easily detect similarities between formal and accurately selected texts. Thus the necessity for our new models.  

\textbf{Direct Quotes/Replies (DQ/DR)}: Collections of $5k$ query tweets, each one paired with $5$ positive candidates (quotes/replies of the query tweets) and $25$ negative candidates (quotes/replies of other tweets). We rank candidates by cosine distance between their embeddings and the embedding of the query tweet. 

\textbf{Co-Quote/Reply (CQ/CR)}: Similar to the previous tasks, we focus on co-quotes/co-replies, i.e., pairs of quotes/replies of the same tweet. These datasets are collections of $5k$ query quotes/replies, each one paired with $5$ positive candidates (quotes/replies of the same tweet) and $25$ negative candidates (quotes/replies of other tweets). We rank candidates by cosine distance between their embeddings and the embedding of the query tweet. 

\subsection{Established benchmarks}

We select two benchmarks from Twitter, PIT dataset and Twitter URL dataset (TURL), and the STS benchmark of formal texts. We pick Pearson correlation coefficient (Pearson's $r$) as metric.

\textbf{PIT-2015 dataset}~\cite{PIT} is a Paraphrase Identification (PI) and Semantic Textual Similarity (SS) task for the Twitter data. It consists in $18762$ sentence pairs annotated with a graded score between 0 (no relation) and 5 (semantic equivalence). We test the models on SS task. 

\textbf{Twitter URL dataset}~\cite{TURL} is the largest human-labeled paraphrase corpus of $51524$ sentence pairs and the first cross-domain benchmarking for automatic paraphrase identification. The data are collected by linking tweets through shared URLs, that are further labeled by human annotators, from 0 to 6. 

\textbf{STS benchmark datasets}~\cite{cer-etal-2017-semeval} is a classical dataset where pairs of formal texts are scored with labels from $0$ to $5$ as semantically similar. It has been widely used to train previous SOTA models, so we do not expect our models trained on informal weak pairs of texts to outperform them. 
However, it is a good indicator of the quality of embeddings and we do expect our models to not deteriorate on accuracy with respect to their initialized versions. 

\subsection{Baselines}

We compare our models with the pre-trained initializations previously described: RoBERTa-base and BERTweet (MEAN pooling of tokens) and S-BERT and S-RoBERTa, pre-trained also on STSb. 

\section{Results and Ablation Study}\label{sec:res}

\begin{table*}
\centering 
\begin{tabular}{l|ccccc||ccc}
\hline 
\textbf{Model}                     & \textbf{DQ}   & \textbf{CQ}   & \textbf{DR}   & \textbf{CR}   & \textbf{Avg} & \textbf{PIT}  & \textbf{TURL} & \textbf{STSb} \\ \hline
RoBERTa-base              & 42.9 & 39.1 & 55.0 & 41.0 & 44.5 & 39.5 & 49.7 & 52.5 \\
BERTweet                  & 46.9 & 42.5 & 56.7 & 44.1 & 47.5 & 38.5 & 48.2 & 48.2 \\
S-BERT                    & 53.7 & 43.9 & 60.5 & 45.4 & 50.9 & 43.8 & \textbf{69.9} & 84.2 \\ 
S-RoBERTa                 & 52.4 & 42.8 & 59.1 & 44.1 & 49.6 & 57.3 & 69.1 & \textbf{84.4}  \\ \hline \hline
Our-RoBERTa-base (all)    & 80.8 & 68.5 & 83.0 & 66.1 & 74.6 & 58.8 & 67.5 & 74.2 \\
Our-BERTweet (all)        & 83.7 & 72.1 & 84.2 & 68.3 & \textbf{77.1} & 66.1 & 67.1 & 72.4 \\ 
Our-S-BERT (all)          & 79.0 & 66.6 & 81.5 & 64.6 & 72.9 & 57.7 & 69.4 & 76.1 \\
Our-S-RoBERTa (all)       & 80.2& 67.8 & 82.6 & 65.6 & 74.0 & 60.1 & 69.0 & 78.9 \\ \hline
Our-RoBERTa (Qt)          & 75.9 & 63.6 & 79.3 & 61.2 & 70.0 & 60.7 & 66.8 & 74.9 \\ 
Our-BERTweet (Qt)         & 80.8 & 68.9 & 81.7 & 65.0 & 74.1 & \textbf{67.4} & 66.0 & 72.4 \\ 
Our-S-BERT (Qt)           & 73.6 & 61.5 & 77.7 & 59.8 & 68.1 & 57.6 & 69.1 & 79.3 \\ 
Our-S-RoBERTa (Qt)        & 74.6 & 62.6 & 78.4 & 60.5 & 69.0 & 58.1 & 68.8 & 80.7 \\ \hline
Our-BERTweet (Co-Qt)      & 80.7 & 70.6 & 80.8 & 65.9 & 74.5 & 63.6 & 64.3 & 70.9 \\  
Our-BERTweet (Rp)         & 81.5 & 68.4 & 82.2 & 65.8 & 74.5 & 63.8 & 67.3 & 72.3 \\
Our-BERTweet (Co-Rp)      & 79.3 & 69.0 & 81.7 & 67.5 & 74.4 & 62.1 & 64.3 & 67.3 \\ \hline

Our-BERTweet-TLoss (Qt) & 67.7 & 60.8 & 71.5 & 56.9 & 64.2 & 53.1 & 43.4 & 44.7  \\ \hline

\hline
\end{tabular}
\caption{$nDCG\times 100$ (novel benchmarks) and Pearson's $r \times 100$ (established benchmarks). We indicate our models with the \textit{Our-} prefix followed by the name of the initialization model, between parentheses the training dataset. If not specified, we use MNLoss. Results are averages of 5 runs.}
\label{tab:res}
\end{table*}

In Table~\ref{tab:res} we show the results of the experiments. 

As expected, we conclude that baseline models perform poorly in the new benchmarks, being trained for different objectives on different data, while \textit{Our-BERTweet (all)} obtains the best performances. 
On established datasets, our training procedure improves the corresponding pre-trained versions. 
The only exception is when our model is initialized from S-BERT and S-RoBERTa and tested on TURL, where we notice a small deterioration of performances (0.5 and 0.1 points respectively) and on STSb-test, since baselines where trained on STSb-train. 
This result proves that our corpora of weakly similar texts are valuable training sets and specific NLI corpora are not necessary to train accurate sentence embeddings. 
We remark that for many non-English languages, models such as S-BERT and S-RoBERTa cannot be trained since datasets such as STSb-train do not exist yet\footnote{Recently, multilingual approaches have been succesfully tested~\cite{reimers-2020-multilingual-sentence-bert}.}. 

The best initialization for novel benchmarks and PIT is BERTweet, being previously unsupervisedly trained on big amounts of similar data, while for TURL and STSb the best initializations are S-BERT and S-RoBERTa respectively. 
MNLoss always produces better results than a simple TripletLoss, since the former compares multiple negative samples for each positive pair, instead of just one as in the latter. 

The training dataset does not largely influence the performance of the model on novel benchmarks, while, on enstablished benchmarks, Qt and Rp are usually better than CoQt and CoRp training datasets. 
However, the concatenation of all datasets (all) used as training set almost always produces better results than when a single dataset is used. 

\begin{figure}[t]
    \centering
    \includegraphics[width=\linewidth]{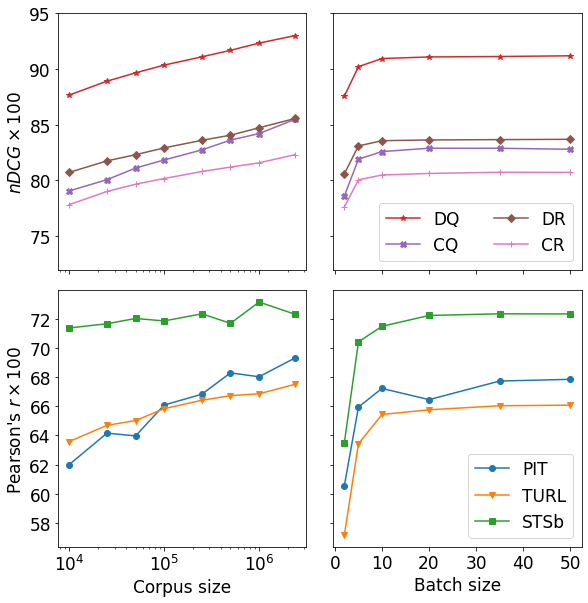}
    \caption{$nDCG\times 100$ and Pearson's $r \times 100$ varying Corpus size (left) and Batch size (right) on Our-BERTweet trained on Quote dataset with MNLoss. Results are averages of 5 runs.}
    \label{fig:sizes}
\end{figure}

Figure~\ref{fig:sizes} (left) shows that performances improve by increasing the corpus size of Qt dataset. 
Since they do not reach a plateau yet, we expect better performances when a wider magnitude of Tweets is collected. 

Figure~\ref{fig:sizes} (right) shows the performance of the same model when varying batch size in MNLoss, i.e., the number of negative samples for each query. 
The performance plateaus at about $10$, setting a sufficient number of negative samples. However, we set it to a higher value because it implies a faster training step. 

\section{Conclusions}
We propose a simple approach to exploit Twitter in building datasets of weak semantically similar texts. Our results prove that exact paraphrases, such as in NLI datasets, are not necessary to train accurate models generating high-quality sentence-embeddings, since models trained on our datasets of \textit{weak} pairs perform well on both established and novel benchmakrs of informal texts. 

The intrinsic relatedness of quotes with quoted texts and replies with the replied texts is particularly useful when building large datasets without human manual effort. 
Thus, we plan to expand the study to other languages spoken in Twitter. Two months of English data are more than enough to build large datasets, but the time window can be easily extended for rarer languages, as today more than $9$ years of data are available to download. 
Finally, we also hypothesize that this approach can be adapted to build high-quality embeddings for text classification tasks. We will extensively explore this on Twitter-related tasks. 
\section{Ethical Considerations}
We generate the training datasets and novel benchmarks starting from the general Twitter Stream collected by the Archive Team Twitter, as described in ~\cref{sec:datasets}. They store data coming from the Twitter Stream and share it in compressed files each month without limits. 
This collection is useful since we can design and perform experiments on Twitter data that are completely reproducible. 
However, it does not honor users' post deletions, account suspensions made by Twitter, or users’ changes from public to private. 
Using Twitter official API to generate a dataset is not a good option for reproducibility since parts of data could be missing due to Twitter Terms of Service. 
We believe that our usage of Twitter Stream Archive is not harmful since we do not collect any delicate information from tweets and users. We download textual data and connections between texts (quotes and replies), and we also remove screen names mentioned in the tweets during the cleaning step. 

However, we agree that Twitter Stream Archive could help malicious and unethical behaviours through inappropriate usage of its data.

\bibliography{biblio}

\begin{thebibliography}{33}
\expandafter\ifx\csname natexlab\endcsname\relax\def\natexlab#1{#1}\fi

\bibitem[{Attardi(2015)}]{Wikiextractor2015}
Giusepppe Attardi. 2015.
\newblock Wikiextractor.
\newblock \url{https://github.com/attardi/wikiextractor}.

\bibitem[{Barbieri et~al.(2020)Barbieri, Camacho-Collados, Espinosa~Anke, and
  Neves}]{barbieri-etal-2020-tweeteval}
Francesco Barbieri, Jose Camacho-Collados, Luis Espinosa~Anke, and Leonardo
  Neves. 2020.
\newblock \href {https://doi.org/10.18653/v1/2020.findings-emnlp.148}
  {{T}weet{E}val: Unified benchmark and comparative evaluation for tweet
  classification}.
\newblock In \emph{Findings of the Association for Computational Linguistics:
  EMNLP 2020}, pages 1644--1650, Online. Association for Computational
  Linguistics.

\bibitem[{Bowman et~al.(2015)Bowman, Angeli, Potts, and
  Manning}]{snli-emnlp2015}
Samuel~R. Bowman, Gabor Angeli, Christopher Potts, and Christopher~D. Manning.
  2015.
\newblock A large annotated corpus for learning natural language inference.
\newblock In \emph{Proceedings of the 2015 Conference on Empirical Methods in
  Natural Language Processing (EMNLP)}. Association for Computational
  Linguistics.

\bibitem[{Carlsson et~al.(2021)Carlsson, Gyllensten, Gogoulou, Hellqvist, and
  Sahlgren}]{carlsson2021semantic}
Fredrik Carlsson, Amaru~Cuba Gyllensten, Evangelia Gogoulou,
  Erik~Ylip{\"a}{\"a} Hellqvist, and Magnus Sahlgren. 2021.
\newblock \href {https://openreview.net/forum?id=Ov_sMNau-PF} {Semantic
  re-tuning with contrastive tension}.
\newblock In \emph{International Conference on Learning Representations}.

\bibitem[{Cer et~al.(2017)Cer, Diab, Agirre, Lopez-Gazpio, and
  Specia}]{cer-etal-2017-semeval}
Daniel Cer, Mona Diab, Eneko Agirre, I{\~n}igo Lopez-Gazpio, and Lucia Specia.
  2017.
\newblock \href {https://doi.org/10.18653/v1/S17-2001} {{S}em{E}val-2017 task
  1: Semantic textual similarity multilingual and crosslingual focused
  evaluation}.
\newblock In \emph{Proceedings of the 11th International Workshop on Semantic
  Evaluation ({S}em{E}val-2017)}, pages 1--14, Vancouver, Canada. Association
  for Computational Linguistics.

\bibitem[{Cer et~al.(2018)Cer, Yang, yi~Kong, Hua, Limtiaco, John, Constant,
  Guajardo-Cespedes, Yuan, Tar, Sung, Strope, and Kurzweil}]{cer2018universal}
Daniel Cer, Yinfei Yang, Sheng yi~Kong, Nan Hua, Nicole Limtiaco, Rhomni~St.
  John, Noah Constant, Mario Guajardo-Cespedes, Steve Yuan, Chris Tar,
  Yun-Hsuan Sung, Brian Strope, and Ray Kurzweil. 2018.
\newblock \href {http://arxiv.org/abs/1803.11175} {Universal sentence encoder}.

\bibitem[{Chen et~al.(2019)Chen, Tang, Wiseman, and
  Gimpel}]{chen-etal-2019-multi-task}
Mingda Chen, Qingming Tang, Sam Wiseman, and Kevin Gimpel. 2019.
\newblock \href {https://doi.org/10.18653/v1/N19-1254} {A multi-task approach
  for disentangling syntax and semantics in sentence representations}.
\newblock In \emph{Proceedings of the 2019 Conference of the North {A}merican
  Chapter of the Association for Computational Linguistics: Human Language
  Technologies, Volume 1 (Long and Short Papers)}, pages 2453--2464,
  Minneapolis, Minnesota. Association for Computational Linguistics.

\bibitem[{Conneau et~al.(2017)Conneau, Kiela, Schwenk, Barrault, and
  Bordes}]{conneau-EtAl:2017:EMNLP2017}
Alexis Conneau, Douwe Kiela, Holger Schwenk, Lo\"{i}c Barrault, and Antoine
  Bordes. 2017.
\newblock \href {https://www.aclweb.org/anthology/D17-1070} {Supervised
  learning of universal sentence representations from natural language
  inference data}.
\newblock In \emph{Proceedings of the 2017 Conference on Empirical Methods in
  Natural Language Processing}, pages 670--680, Copenhagen, Denmark.
  Association for Computational Linguistics.

\bibitem[{Creutz(2018)}]{creutz-2018-open}
Mathias Creutz. 2018.
\newblock \href {https://www.aclweb.org/anthology/L18-1218} {Open subtitles
  paraphrase corpus for six languages}.
\newblock In \emph{Proceedings of the Eleventh International Conference on
  Language Resources and Evaluation ({LREC} 2018)}, Miyazaki, Japan. European
  Language Resources Association (ELRA).

\bibitem[{Devlin et~al.(2019)Devlin, Chang, Lee, and
  Toutanova}]{devlin-etal-2019-bert}
Jacob Devlin, Ming-Wei Chang, Kenton Lee, and Kristina Toutanova. 2019.
\newblock \href {https://doi.org/10.18653/v1/N19-1423} {{BERT}: Pre-training of
  deep bidirectional transformers for language understanding}.
\newblock In \emph{Proceedings of the 2019 Conference of the North {A}merican
  Chapter of the Association for Computational Linguistics: Human Language
  Technologies, Volume 1 (Long and Short Papers)}, pages 4171--4186,
  Minneapolis, Minnesota. Association for Computational Linguistics.

\bibitem[{Dolan and Brockett(2005)}]{dolan-brockett-2005-automatically}
William~B. Dolan and Chris Brockett. 2005.
\newblock \href {https://www.aclweb.org/anthology/I05-5002} {Automatically
  constructing a corpus of sentential paraphrases}.
\newblock In \emph{Proceedings of the Third International Workshop on
  Paraphrasing ({IWP}2005)}.

\bibitem[{Du et~al.(2021)Du, Grave, Gunel, Chaudhary, Celebi, Auli, Stoyanov,
  and Conneau}]{du2020self}
Jingfei Du, Edouard Grave, Beliz Gunel, Vishrav Chaudhary, Onur Celebi, Michael
  Auli, Veselin Stoyanov, and Alexis Conneau. 2021.
\newblock \href {https://doi.org/10.18653/v1/2021.naacl-main.426}
  {Self-training improves pre-training for natural language understanding}.
\newblock In \emph{Proceedings of the 2021 Conference of the North American
  Chapter of the Association for Computational Linguistics: Human Language
  Technologies}, pages 5408--5418, Online. Association for Computational
  Linguistics.

\bibitem[{Giorgi et~al.(2020)Giorgi, Nitski, Bader, and
  Wang}]{giorgi2020declutr}
John~M. Giorgi, Osvald Nitski, Gary~D. Bader, and Bo~Wang. 2020.
\newblock \href {http://arxiv.org/abs/2006.03659} {Declutr: Deep contrastive
  learning for unsupervised textual representations}.

\bibitem[{Gokaslan and Cohen(2019)}]{Gokaslan2019OpenWeb}
Aaron Gokaslan and Vanya Cohen. 2019.
\newblock Openwebtext corpus.
\newblock \small{\url{http://Skylion007.github.io/OpenWebTextCorpus}}.

\bibitem[{Henderson et~al.(2017)Henderson, Al-Rfou, Strope, hsuan Sung,
  Lukács, Guo, Kumar, Miklos, and Kurzweil}]{loss_multiple}
Matthew Henderson, Rami Al-Rfou, Brian Strope, Yun hsuan Sung, László
  Lukács, Ruiqi Guo, Sanjiv Kumar, Balint Miklos, and Ray Kurzweil. 2017.
\newblock Efficient natural language response suggestion for smart reply.
\newblock \emph{ArXiv e-prints}.

\bibitem[{Huang et~al.(2021)Huang, Huang, and Chang}]{huang2021disentangling}
James~Y. Huang, Kuan-Hao Huang, and Kai-Wei Chang. 2021.
\newblock \href {https://doi.org/10.18653/v1/2021.naacl-main.108}
  {Disentangling semantics and syntax in sentence embeddings with pre-trained
  language models}.
\newblock In \emph{Proceedings of the 2021 Conference of the North American
  Chapter of the Association for Computational Linguistics: Human Language
  Technologies}, pages 1372--1379, Online. Association for Computational
  Linguistics.

\bibitem[{J\"{a}rvelin and Kek\"{a}l\"{a}inen(2002)}]{nDCG}
Kalervo J\"{a}rvelin and Jaana Kek\"{a}l\"{a}inen. 2002.
\newblock \href {https://doi.org/10.1145/582415.582418} {Cumulated gain-based
  evaluation of ir techniques}.
\newblock \emph{ACM Trans. Inf. Syst.}, 20(4):422–446.

\bibitem[{Lan et~al.(2017)Lan, Qiu, He, and Xu}]{TURL}
Wuwei Lan, Siyu Qiu, Hua He, and Wei Xu. 2017.
\newblock \href {http://aclweb.org/anthology/D17-1127} {A continuously growing
  dataset of sentential paraphrases}.
\newblock In \emph{Proceedings of The 2017 Conference on Empirical Methods on
  Natural Language Processing (EMNLP)}, pages 1235--1245. Association for
  Computational Linguistics.

\bibitem[{Li et~al.(2020)Li, Zhou, He, Wang, Yang, and
  Li}]{li-etal-2020-sentence}
Bohan Li, Hao Zhou, Junxian He, Mingxuan Wang, Yiming Yang, and Lei Li. 2020.
\newblock \href {https://doi.org/10.18653/v1/2020.emnlp-main.733} {On the
  sentence embeddings from pre-trained language models}.
\newblock In \emph{Proceedings of the 2020 Conference on Empirical Methods in
  Natural Language Processing (EMNLP)}, pages 9119--9130, Online. Association
  for Computational Linguistics.

\bibitem[{{Liu} et~al.(2019){Liu}, {Ott}, {Goyal}, {Du}, {Joshi}, {Chen},
  {Levy}, {Lewis}, {Zettlemoyer}, and {Stoyanov}}]{roberta}
Yinhan {Liu}, Myle {Ott}, Naman {Goyal}, Jingfei {Du}, Mandar {Joshi}, Danqi
  {Chen}, Omer {Levy}, Mike {Lewis}, Luke {Zettlemoyer}, and Veselin
  {Stoyanov}. 2019.
\newblock \href {http://arxiv.org/abs/1907.11692} {{RoBERTa: A Robustly
  Optimized BERT Pretraining Approach}}.
\newblock \emph{arXiv e-prints}, page arXiv:1907.11692.

\bibitem[{Logeswaran and Lee(2018)}]{logeswaran2018an}
Lajanugen Logeswaran and Honglak Lee. 2018.
\newblock \href {https://openreview.net/forum?id=rJvJXZb0W} {An efficient
  framework for learning sentence representations}.
\newblock In \emph{International Conference on Learning Representations}.

\bibitem[{Mikolov et~al.(2013)Mikolov, Sutskever, Chen, Corrado, and
  Dean}]{NIPS2013_word2vec}
Tomas Mikolov, Ilya Sutskever, Kai Chen, Greg~S Corrado, and Jeff Dean. 2013.
\newblock \href
  {https://proceedings.neurips.cc/paper/2013/file/9aa42b31882ec039965f3c4923ce901b-Paper.pdf}
  {Distributed representations of words and phrases and their
  compositionality}.
\newblock In \emph{Advances in Neural Information Processing Systems},
  volume~26. Curran Associates, Inc.

\bibitem[{Nguyen et~al.(2020)Nguyen, Vu, and
  Tuan~Nguyen}]{nguyen-etal-2020-bertweet}
Dat~Quoc Nguyen, Thanh Vu, and Anh Tuan~Nguyen. 2020.
\newblock \href {https://doi.org/10.18653/v1/2020.emnlp-demos.2} {{BERT}weet: A
  pre-trained language model for {E}nglish tweets}.
\newblock In \emph{Proceedings of the 2020 Conference on Empirical Methods in
  Natural Language Processing: System Demonstrations}, pages 9--14, Online.
  Association for Computational Linguistics.

\bibitem[{Pennington et~al.(2014)Pennington, Socher, and
  Manning}]{pennington2014glove}
Jeffrey Pennington, Richard Socher, and Christopher~D. Manning. 2014.
\newblock \href {http://www.aclweb.org/anthology/D14-1162} {Glove: Global
  vectors for word representation}.
\newblock In \emph{Empirical Methods in Natural Language Processing (EMNLP)},
  pages 1532--1543.

\bibitem[{Reimers and Gurevych(2019)}]{reimers-2019-sentence-bert}
Nils Reimers and Iryna Gurevych. 2019.
\newblock \href {https://arxiv.org/abs/1908.10084} {Sentence-bert: Sentence
  embeddings using siamese bert-networks}.
\newblock In \emph{Proceedings of the 2019 Conference on Empirical Methods in
  Natural Language Processing}. Association for Computational Linguistics.

\bibitem[{Reimers and Gurevych(2020)}]{reimers-2020-multilingual-sentence-bert}
Nils Reimers and Iryna Gurevych. 2020.
\newblock \href {https://arxiv.org/abs/2004.09813} {Making monolingual sentence
  embeddings multilingual using knowledge distillation}.
\newblock In \emph{Proceedings of the 2020 Conference on Empirical Methods in
  Natural Language Processing}. Association for Computational Linguistics.

\bibitem[{Vaswani et~al.(2017)Vaswani, Shazeer, Parmar, Uszkoreit, Jones,
  Gomez, Kaiser, and Polosukhin}]{attention}
Ashish Vaswani, Noam Shazeer, Niki Parmar, Jakob Uszkoreit, Llion Jones,
  Aidan~N. Gomez, Lukasz Kaiser, and Illia Polosukhin. 2017.
\newblock \href {https://arxiv.org/pdf/1706.03762.pdf} {Attention is all you
  need}.

\bibitem[{Wang et~al.(2021)Wang, Reimers, and Gurevych}]{wang2021tsdae}
Kexin Wang, Nils Reimers, and Iryna Gurevych. 2021.
\newblock \href {http://arxiv.org/abs/2104.06979} {Tsdae: Using
  transformer-based sequential denoising auto-encoder for unsupervised sentence
  embedding learning}.

\bibitem[{Wieting and Gimpel(2018)}]{wieting-gimpel-2018-paranmt}
John Wieting and Kevin Gimpel. 2018.
\newblock \href {https://doi.org/10.18653/v1/P18-1042} {{P}ara{NMT}-50{M}:
  Pushing the limits of paraphrastic sentence embeddings with millions of
  machine translations}.
\newblock In \emph{Proceedings of the 56th Annual Meeting of the Association
  for Computational Linguistics (Volume 1: Long Papers)}, pages 451--462,
  Melbourne, Australia. Association for Computational Linguistics.

\bibitem[{Wieting et~al.(2020)Wieting, Neubig, and
  Berg-Kirkpatrick}]{wieting-etal-2020-bilingual}
John Wieting, Graham Neubig, and Taylor Berg-Kirkpatrick. 2020.
\newblock \href {https://doi.org/10.18653/v1/2020.emnlp-main.122} {A bilingual
  generative transformer for semantic sentence embedding}.
\newblock In \emph{Proceedings of the 2020 Conference on Empirical Methods in
  Natural Language Processing (EMNLP)}, pages 1581--1594, Online. Association
  for Computational Linguistics.

\bibitem[{Williams et~al.(2018)Williams, Nangia, and
  Bowman}]{williams-etal-2018-broad}
Adina Williams, Nikita Nangia, and Samuel Bowman. 2018.
\newblock \href {https://doi.org/10.18653/v1/N18-1101} {A broad-coverage
  challenge corpus for sentence understanding through inference}.
\newblock In \emph{Proceedings of the 2018 Conference of the North {A}merican
  Chapter of the Association for Computational Linguistics: Human Language
  Technologies, Volume 1 (Long Papers)}, pages 1112--1122, New Orleans,
  Louisiana. Association for Computational Linguistics.

\bibitem[{Wolf et~al.(2020)Wolf, Debut, Sanh, Chaumond, Delangue, Moi, Cistac,
  Rault, Louf, Funtowicz, Davison, Shleifer, von Platen, Ma, Jernite, Plu, Xu,
  Le~Scao, Gugger, Drame, Lhoest, and Rush}]{wolf-etal-2020-transformers}
Thomas Wolf, Lysandre Debut, Victor Sanh, Julien Chaumond, Clement Delangue,
  Anthony Moi, Pierric Cistac, Tim Rault, Remi Louf, Morgan Funtowicz, Joe
  Davison, Sam Shleifer, Patrick von Platen, Clara Ma, Yacine Jernite, Julien
  Plu, Canwen Xu, Teven Le~Scao, Sylvain Gugger, Mariama Drame, Quentin Lhoest,
  and Alexander Rush. 2020.
\newblock \href {https://doi.org/10.18653/v1/2020.emnlp-demos.6} {Transformers:
  State-of-the-art natural language processing}.
\newblock In \emph{Proceedings of the 2020 Conference on Empirical Methods in
  Natural Language Processing: System Demonstrations}, pages 38--45, Online.
  Association for Computational Linguistics.

\bibitem[{Xu et~al.(2015)Xu, Callison-Burch, and Dolan}]{PIT}
Wei Xu, Chris Callison-Burch, and Bill Dolan. 2015.
\newblock \href {https://doi.org/10.18653/v1/S15-2001} {{S}em{E}val-2015 task
  1: Paraphrase and semantic similarity in {T}witter ({PIT})}.
\newblock In \emph{Proceedings of the 9th International Workshop on Semantic
  Evaluation ({S}em{E}val 2015)}, pages 1--11, Denver, Colorado. Association
  for Computational Linguistics.

\end{thebibliography}
\bibliographystyle{acl_natbib}

\section*{Appendix}
\begin{figure*}[]
    \centering
    \includegraphics[width=\linewidth]{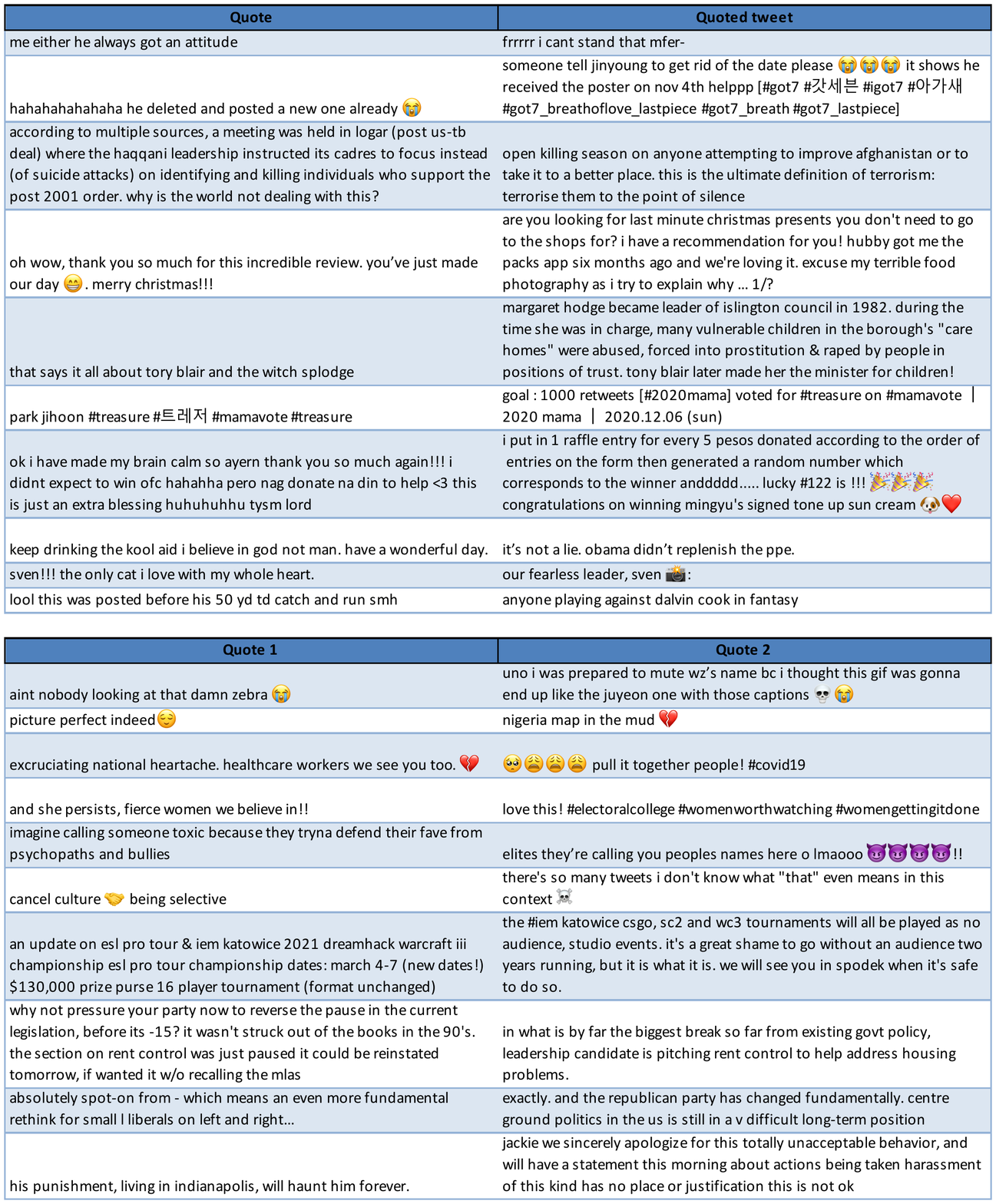}
    \caption{Examples of pairs of texts from Qt (top) and CoQt (bottom) datasets.}
    \label{fig:samples1}
\end{figure*}

\begin{figure*}[]
    \centering
    \includegraphics[width=\linewidth]{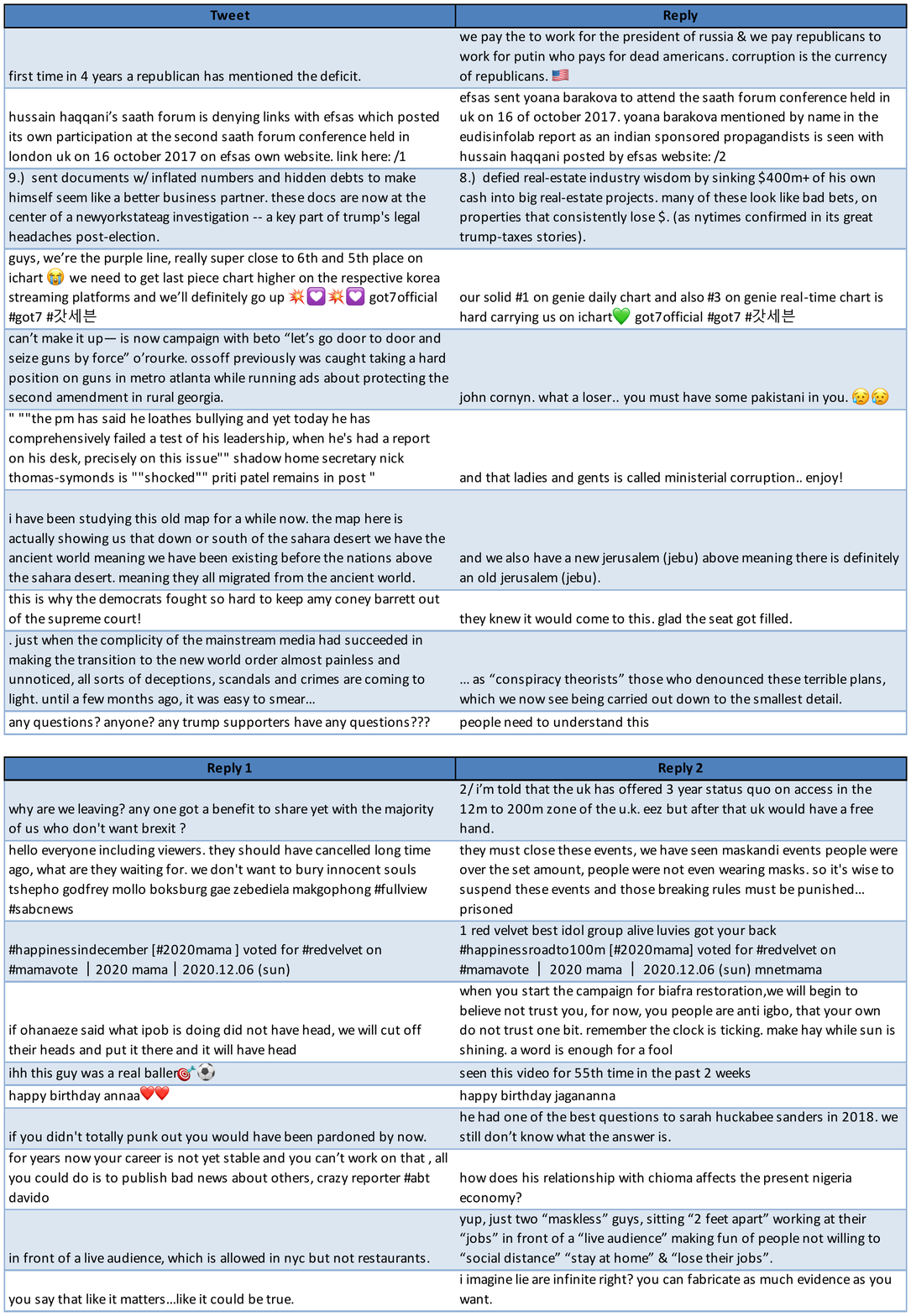}
    \caption{Examples of pairs of texts from Rp (top) and CoRp (bottom) datasets.}
    \label{fig:samples2}
\end{figure*}

We show randomly selected examples of training data from Qt and CoQt datasets in Figure~\ref{fig:samples1}, and from Rp and CoRp datasets in Figure~\ref{fig:samples2}.

\end{document}